\documentclass[runningheads]{llncs}
\usepackage[T1]{fontenc}
\usepackage{fmtcount}
\usepackage[colorlinks,linkcolor=blue]{hyperref}
\usepackage{graphicx,verbatim}
\usepackage{amssymb}
\usepackage{amsmath}
\usepackage{amssymb}
\usepackage{multirow}
\usepackage{tabularx}
\usepackage{url}
\usepackage[table,xcdraw]{xcolor}
\usepackage{amsmath}
\usepackage{xcolor}
\usepackage{graphicx} 
\usepackage{booktabs}
\usepackage{mathrsfs}
\usepackage{makecell} 
\usepackage{caption} 
\usepackage[misc]{ifsym}

\begin{document}
\title{MReg: A Novel Regression Model with MoE-based Video Feature Mining for Mitral Regurgitation Diagnosis}
\author{
Zhe Liu\inst{1,2}\thanks{Zhe Liu, Yuhao Huang and Lian Liu contribute equally to this work.\\ 
Corresponding email: \email{gzhongshan1986@163.com} and  \email{xinyang@szu.edu.cn}}\and
Yuhao Huang\inst{1,2\star}\and 
Lian Liu\inst{1,2,3\star}\and 
Chengrui Zhang\inst{4}\and 
Haotian Lin\inst{3}\and\ 
Tong Han\inst{1,2}\and 
Zhiyuan Zhu\inst{1,2}\and 
Yanlin Chen\inst{1,2}\and 
Yuerui Chen\inst{5}\and 
Dong Ni\inst{2}\and 
Zhongshan Gou\inst{6}\textsuperscript{(\Letter)}\and 
Xin Yang\inst{1,2}\textsuperscript{(\Letter)} 
}
\authorrunning{Liu et al.}
\titlerunning{MoE Video Mining for MR}

%
\institute{\textsuperscript{$1$}National-Regional Key Technology Engineering Laboratory for Medical Ultrasound, School of Biomedical Engineering, Medical School,
Shenzhen University, China\\
\textsuperscript{$2$}Medical Ultrasound Image Computing (MUSIC) Lab, School of Biomedical Engineering, Medical School, Shenzhen University, China\\
\textsuperscript{$3$}Shenzhen RayShape Medical Technology Co., Ltd, China\\
\textsuperscript{$4$}School of Physics and Astronomy, The University of Edinburgh, United Kingdom\\
\textsuperscript{$5$}School of Future Technology, South China University of Technology, China\\
\textsuperscript{$6$}Center for Cardiovascular Disease, The Affiliated Suzhou Hospital of Nanjing Medical University, China\\
}
\maketitle              
\begin{abstract}
Color Doppler echocardiography is a crucial tool for diagnosing mitral regurgitation (MR).
Recent studies have explored intelligent methods for MR diagnosis to minimize user dependence and improve accuracy.
However, these approaches often fail to align with clinical workflow and may lead to suboptimal accuracy and interpretability.
In this study, we introduce an automated MR diagnosis model (\textit{MReg}) developed on the 4-chamber cardiac color Doppler echocardiography video (A4C-CDV).
It follows comprehensive feature mining strategies to detect MR and assess its severity, considering clinical realities.
Our contribution is threefold.
First, we formulate the MR diagnosis as a regression task to capture the continuity and ordinal relationships between categories.
Second, we design a feature selection and amplification mechanism to imitate the sonographer's diagnostic logic for accurate MR grading.
Third, inspired by the Mixture-of-Experts concept, we introduce a feature summary module to extract the category-level features, enhancing the representational capacity for more accurate grading.
We trained and evaluated our proposed $MReg$ on a large in-house A4C-CDV dataset comprising 1868 cases with three graded regurgitation labels. 
Compared to other weakly supervised video anomaly detection and supervised classification methods, $MReg$ demonstrated superior performance in MR diagnosis.
Our code is available at: \url{https://github.com/cskdstz/MReg}.
\keywords{Mitral regurgitation \and Color Doppler Echocardiography \and Feature Mining \and Mixture-of-Experts}
\end{abstract}
%
%
%
\section{Introduction}
Mitral regurgitation (MR) is a prevalent valvular heart disease, with precise severity assessment essential for treatment and prognosis.
Color Doppler echocardiography is the most common tool in MR diagnosis due to its non-invasive nature, offering a clear visualization of blood flow to sonographers for disease assessment.
However, MR diagnosis heavily relies on the sonographer's experience, introducing subjectivity, increasing the risk of misdiagnosis and limiting the feasibility of large-scale screening. 
Hence, developing an automated system for MR diagnosis is crucial to improving diagnostic objectivity and accuracy.

Deep learning has been widely applied in the analysis of valvular heart disease, including structural segmentation, parametric measurements~\cite{chandra2020mitral,herz2021segmentation,li2023echoefnet}, and disease detection~\cite{wahlang2021deep,yang2022automated,cheng2022revealing,liu2024mitral}, e.g., mitral/aortic/tricuspid regurgitation.
However, these methods were not specifically designed for MR diagnosis, which may lead to suboptimal accuracy. 
Recently, Sadeghpour et al.~\cite{sadeghpour2025automated} developed a multiparametric machine-learning method to measure MR-related parameters for grading based on images.
However, their strategies heavily relied on keyframe selection, ignoring crucial dynamic information for accurate assessment.

\begin{figure}[!t]
\centering
\includegraphics[width=0.81\linewidth]{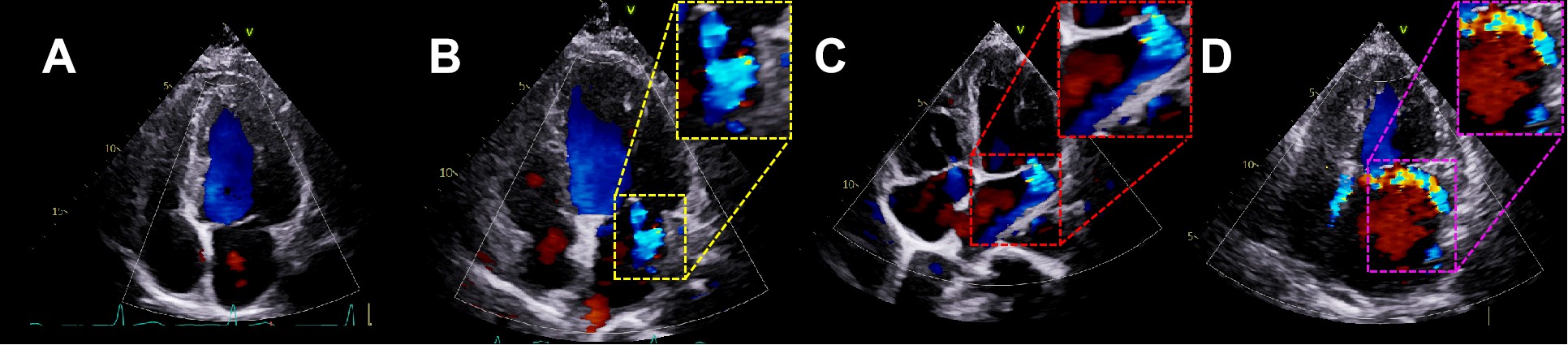}
\caption{Visualizations of normal (A) and various MR cases: (B) mild MR, (C)-(D) eccentric moderate and severe MR. Dashed boxes show the regurgitation jets.}
\label{fig:MR_data}
\end{figure}

Two studies~\cite{long2024deep,vrudhula2024high} used the naive video-based convolutional neural network model to classify MR.
Although effective, they utilized conventional video classification models that were not specifically optimized for MR features. 
Thus, these methods may struggle with complex cases, such as multi-jet or eccentric regurgitation. 
Figure~\ref{fig:MR_data} shows that moderate and severe MR with eccentricity have similar regurgitant areas to mild MR, which may confuse the above methods and lead to inaccurate prediction.
Moreover, data imbalances can lead to significant variations in diagnostic accuracy across categories, e.g., higher misdiagnosis rates for moderate MR~\cite{sadeghpour2025automated}. Hence, an automated framework to address the above challenges is highly desirable for accurate and robust MR diagnosis.

In this paper, we develop a novel regression-based framework named \textit{MReg} to predict normal (Grade 0), mild MR (Grade 1), and moderate-severe MR (Grade
2), using A4C-CDV.
We combine moderate and severe MR, as both require clinical intervention, unlike mild MR requiring only periodic screening. 
Based on three feature mining methods, our highlights are: 
\textbf{(1)} To our knowledge, this is the first end-to-end video-based MR-specific qualitative diagnostic model formulated as a regression task to capture category relationships.
\textbf{(2)} We introduce $feature$ $selection$ and $amplification$ designs to mimic the clinical process, driven by multiple instance learning (MIL). 
\textbf{(3)} We propose an Mixture-of-Experts (MoE)-based $feature$ $summary$ module to achieve category-level feature decoupling for handling complex cases.
\textbf{(4)} Extensive experiments show that our method significantly outperforms existing strong video-based methods.

\section{Methodology}
\begin{figure}[!t]
\centering
\includegraphics[width=1.0\linewidth]{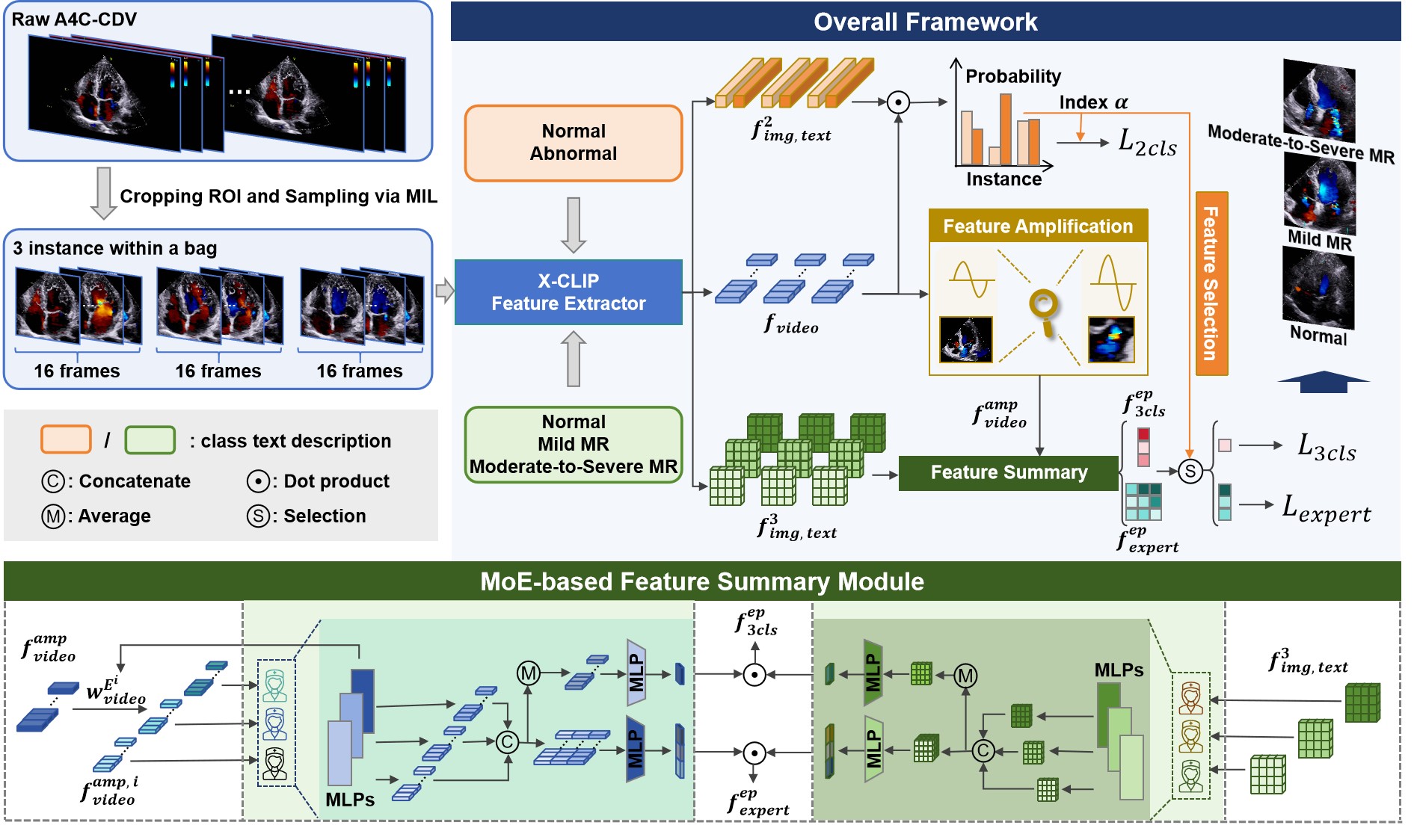}
\caption{Overview of our proposed method.} \label{fig:framewaork}
\end{figure}

Figure~\ref{fig:framewaork} shows the overview of our proposed framework.
Videos are first cropped to retain the region of interest (ROI), and sampled via MIL into three 16-frame instances.
The MIL-based operation is designed to facilitate the subsequent feature selection to use the most representative clip for diagnosis, avoiding interference from others.
Then, X-CLIP~\cite{ni2022expanding} is used to extract video features $f_{video} \in R^{3 \times 16 \times 512}$, and two/three category text features: $f_{img, text}^{2} \in R^{3 \times 2 \times 512}$ and $f_{img, text}^{3} \in R^{3 \times 16 \times 196 \times 3 \times 512}$, respectively.
Last, these features will be input into the following framework to perform efficient and accurate MR diagnosis.

Notably, previous methods often treat MR diagnosis as a classification task. 
However, this approach primarily suffers from the drawback that discrete category labels in classification scenarios may fail to capture the continuity and ordinal relationships between categories.
Besides, the cross-entropy loss may further aggravate the learning of relative category relations, potentially breaking the learning stability.
Hence, given the continuous spectrum of MR severity, we reformulate the MR diagnosis as a regression task and propose $MReg$ to effectively model category relationships.
Below are the method details, including three feature mining methods: feature selection, amplification, and summary.

\subsection{Feature Selection and Amplification for Coarse MR Regression}

Clinically, A4C-CDV-based MR diagnosis mainly contains two steps. 
First, sonographers determine the presence of MR and identify the cardiac cycle with the most severe regurgitation for severity assessment.
Second, sonographers combine A4C-CDV local amplification video to observe the regurgitation area in more detail to achieve an accurate diagnosis. 
Inspired by this, we design a two-stage framework to match the clinical behavioral logic of MR analysis.

At stage I (i.e., \textbf{Step 1}), we follow the X-CLIP mechanism to determine the presence of MR and select the instance index $\alpha=argmax(f_{2cls\_{out}} \allowbreak  \left[:, 1\right])$ with the highest MR probability for subsequent diagnosis.
The focal loss~\cite{lin2017focal} for binary classification $L_{2cls}=FocalLoss(SoftMax(S(f_{2cls\_{out}}; \alpha)), \overline{Y}_2)$ supervises this process. 
$f_{2cls\_{out}}$ represents the model output, $S(\cdot)$ represents the selection of corresponding instance features according to $\alpha$, and $\overline{Y}_2$ represents the 2-category label, i.e., normal or MR.
At stage II (i.e., \textbf{Step 2}), $f_{video}$ and $f_{img, text}^{3}$ will be used for MR diagnosis based on $\alpha$. Here, we propose a feature amplification mechanism to amplify the features, which not only retains the complete cardiac motion information in A4C-CDV, but also amplifies the regurgitation feature information to facilitate MR diagnosis. The amplified feature $f_{video}^{amp}$ is: 

\begin{equation}
f_{video}^{amp}=\mathcal{N}(f_{video} + \beta \times Conv(\| f_{video} \|_2)),
\end{equation}
where $\mathcal{N}(\cdot)$ represents the feature normalization on the time and instance dimension to facilitate the subsequent calculation.
$\beta$ is the amplification coefficient. 
$Conv(\cdot)$ represents a one-dimensional convolution to select the amplified features that are MR-relevant, and $\|\cdot\|_2$ indicates $L_2$ norm.
The combination of amplified and original features not only retains the information extracted from the complete heart, but also adds effective local amplified information.
It is noted that both features are extracted from the same video for convenient combination.
Last, we use MSE loss~\cite{ren2022balanced} for model optimization:
\begin{equation}
L_{3cls}=MSELoss(S(\mathcal{F}(\\f_{video}^{amp}, f_{img, text}^3); \alpha), \overline{Y}_3),
\label{eq:L3}
\end{equation}
where $\mathcal{F}(\cdot)$ denotes dot product fusion after feature dimension transformation, and $\overline{Y}_3$ denote the 3-category label. 
We transform the model output into three categories based on thresholds $thre_1$ and $thre_2$.
In this case, even if the regression model uses discrete labels, the continuity and order relationships of different categories can still be learned to enhance MR diagnosis.

\subsection{Enhanced Regression via MoE-based Feature Summary Module}
While the above feature selection and amplification can mimic the clinical workflow and yield good results, the complexity of the MR diagnostic mechanism still leads to unsatisfactory accuracy.
Specifically, MR severity is influenced by various factors, such as regurgitation jet area, duration of regurgitation, and degree of eccentricity of regurgitation jet, and MR of varying severity typically contains different characteristics.
Inspired by these clinical facts, we introduce the MoE mechanism and built the feature summary module with three experts for $f_{img, text}^{3}$ and $f_{video}^{amp}$, respectively.
Then, the proposed module aims to summarize MR features in different categories. Since $f_{img, text}^{3}$ has features of three categories and $f_{video}^{amp}$ does not, we map $f_{video}^{amp}$ into three directions to obtain $f_{video}^{amp, i}$, which is then summarized it by different experts:
\begin{equation}
w_{video}^{E^{i}}=SoftMax(\| W_{video}^i \|_F);
f_{video}^{amp, i}=f_{video}^{amp} \odot w_{video}^{E^{i}},
\end{equation}
where $\| \cdot \|_F$ is the Frobenius norm~\cite{bottcher2008frobenius}, and $W_{video}^i \in R^{512 \times 512}$ means the weight of each video expert. \textit{SoftMax} function highlights relevant features by augmenting the weight of the most relevant expert and reducing the influence of irrelevant ones. 
The regression loss function of stage II is updated from \eqref{eq:L3} to $L_{3cls}=MSELoss(S(f_{3cls}^{ep}; \alpha), \overline{Y}_3)$, where $f_{3cls}^{ep}$ denotes the model output after mixing three experts in the feature summary module.
Additionally, we designed a focal loss based function to optimize the expert learning:
\begin{equation}
L_{expert}=FocalLoss(argmax(S(f_{expert}^{ep}; \alpha)), \overline{Y}_3),
\end{equation}
where $f_{expert}^{ep}$ denotes the model output without mixing three experts. 
Therefore, with the smoothing $L_{smooth}$ and sparsity losses $L_{sparsity}$, the total loss function can be written as: $L_{total}=L_{2cls}+L_{3cls}+L_{expert}+\lambda_1 L_{smooth}+\lambda_2 L_{sparsity}$, where $\lambda_1$ and $\lambda_2$ are two weight hyperparameters. 

\section{Experimental Results}
\subsubsection{Dataset and Implementations.} 

The dataset comprised 1,868 A4C-CDV cases, categorized into Grade 0 (normal, 965), Grade 1 (mild MR, 677), and Grade 2 (moderate-severe MR, 226) by an experienced sonographer.
Then, cases were randomly divided for training (450/296/103), validation (112/74/25), and testing (403/307/98).
All frames in the videos were resized to 224$\times$224 after ROI detection and MIL sampling. 
ROI is a rectangular area covering all cardiac chambers, obtained using YOLOv8~\cite{Terven2023comprehensive} pre-trained on our MR dataset with ROI labels annotated using Pair software~\cite{liang2022sketch}. 
The MIL sampling uses 3×16-frame non-overlapping clips per video. The data acquisition protocol ensured that each consecutive 16-frame clip encompassed more than 1 cardiac cycle.

All experiments were implemented in $PyTorch$ (version 2.0.1) with two NVIDIA 3090 GPUs with 24G memory each.
The training was optimized by AdamW in 100 epochs, with batch size=1 and learning rate=1e-5.  
We used X-CLIP pretrained on Kinetics-400~\cite{carreira2017quo} for initialization. 
The amplification coefficient $\beta$ was 2, and the thresholds $thre_1$\&$thre_2$ were 0.5\&1.5, respectively. 
In the loss function, we set $\lambda_1$=0.01 and $\lambda_2$=0.001. 
We employed five evaluation metrics: Accuracy (\%), Recall (\%), Precision (\%), Specificity (\%), and F1-score (\%).
The model with the highest validation accuracy will be selected for evaluation.

\subsubsection{Quantitative and Qualitative Analysis.} We validated the effectiveness of the proposed $MReg$ through comprehensive comparative experiments, ablation studies, and diverse visualizations. 
We also performed Chi-square test results between $MReg$ and other methods for significance analysis.

Table~\ref{tab:comparison of methods} compares $MReg$ with supervised video classification (SVC) and weakly supervised video anomaly detection (WS-VAD) methods. 
For SVC methods, we compare $MReg$ with three widely used approaches, including ViVit~\cite{arnab2021vivit},  ActionCLIP~\cite{wang2021actionclip} and VideoSwin~\cite{liu2022video}.
WS-VAD-based methods are included since they share a similar mechanism with $MReg$, learning frame-level anomalies using video-level labels.
However, since $MReg$ not only detects anomalies (MR) but also categorizes them, we specifically consider AnomalyCLIP~\cite{zanella2024delving} with classification ability for comparison. 
The results demonstrate that $MReg$ achieves state-of-the-art performance compared to all competing methods. 
Notably, WS-VAD methods perform significantly worse than SVC methods on all metrics.
However, despite sharing a similar mechanism with WS-VAD, $MReg$ significantly surpasses all SVC methods, further highlighting its superiority.

\begin{table}[!t]
  \caption{Method Comparison. The best results are shown in bold.}
  \setlength{\tabcolsep}{1.5mm}
    \begin{tabular}{c|cccccc}
    \Xhline{1.0px}
     SW-VAD Methods & \multicolumn{1}{l}{Accuracy} & \multicolumn{1}{l}{Recall} & \multicolumn{1}{l}{Precision} & \multicolumn{1}{l}{Specificity} & \multicolumn{1}{l}{F1-score} & \multicolumn{1}{l}{P-Value}\\
    \hline
    AnomalyCLIP~\cite{zanella2024delving} & 65.35 & 60.78 & 60.08 & 60.12 & 80.50 & < .001\\
    \hline
    SVC Methods & \multicolumn{1}{l}{Accuracy} & \multicolumn{1}{l}{Recall} & \multicolumn{1}{l}{Precision} & \multicolumn{1}{l}{Specificity} & \multicolumn{1}{l}{F1-score} & \multicolumn{1}{l}{P-Value}\\
    \hline
    ViVit~\cite{arnab2021vivit} & 72.28 & 68.43 & 76.03 & 69.74 & 82.72 & < .001\\
    ActionCLIP~\cite{wang2021actionclip} & 82.67 & 82.36 & 81.28 & 81.47 & 90.86 & < .001 \\
    Video-Swin~\cite{liu2022video} & 86.76 & 85.89 & 82.94 & 84.22 & 93.18 & < .001\\
    $MReg$  & \textbf{89.36} & \textbf{85.93} & \textbf{86.83} & \textbf{86.36} & \textbf{94.28} &  -\\
    \Xhline{1.0px}
    \end{tabular}%
  \label{tab:comparison of methods}%
\end{table}%

\begin{table}[!t]
  \centering
      \small
  \caption{MR diagnostic performance of the model with different modules.}
  \resizebox{\textwidth}{!}{%
    \begin{tabular}{c|cccccc}
    \Xhline{1.0px}
    Model & \multicolumn{1}{l}{Accuracy} & \multicolumn{1}{l}{Recall} & \multicolumn{1}{l}{Precision} & \multicolumn{1}{l}{Specificity} & \multicolumn{1}{l}{F1-score} & \multicolumn{1}{l}{P-Value}\\
    \hline
    Baseline & 82.67 & 75.57 & 81.31 & 77.73 & 90.09 & < .001\\
    Baseline + FS & 87.38 & 83.45 & 84.42 & 83.90 & 93.25 & < .001\\
    Baseline + FS + Amp & 88.74 & 84.87 & 85.60 & 85.23 & 93.98 & < .001 \\
    Baseline* + FS + Amp & 86.88 & 82.99 & 86.83 & 84.43 & 92.83 & < .001\\
    Baseline + FS + Amp + MoE & 88.37 & 85.37 & 86.68 & 85.80 & 93.90 & < .001 \\
    $MReg$** & 81.06 & 74.87 & 81.46 & 77.13 & 88.59 & < .001 \\
    $MReg$  & \textbf{89.36} & \textbf{85.93} & \textbf{86.83} & \textbf{86.36} & \textbf{94.28} & - \\
    \Xhline{1.0px}
    \end{tabular}}%
    \raggedright
    \par\footnotesize 
    * Corresponds to the $classification$ model.
    
    ** Indicates the removal of FS in $MReg$ and randomly selecting instances.
  \label{tab:ablation experiment}%
\end{table}%

\begin{figure}[!t]
\centering
\includegraphics[width=1.0\linewidth]{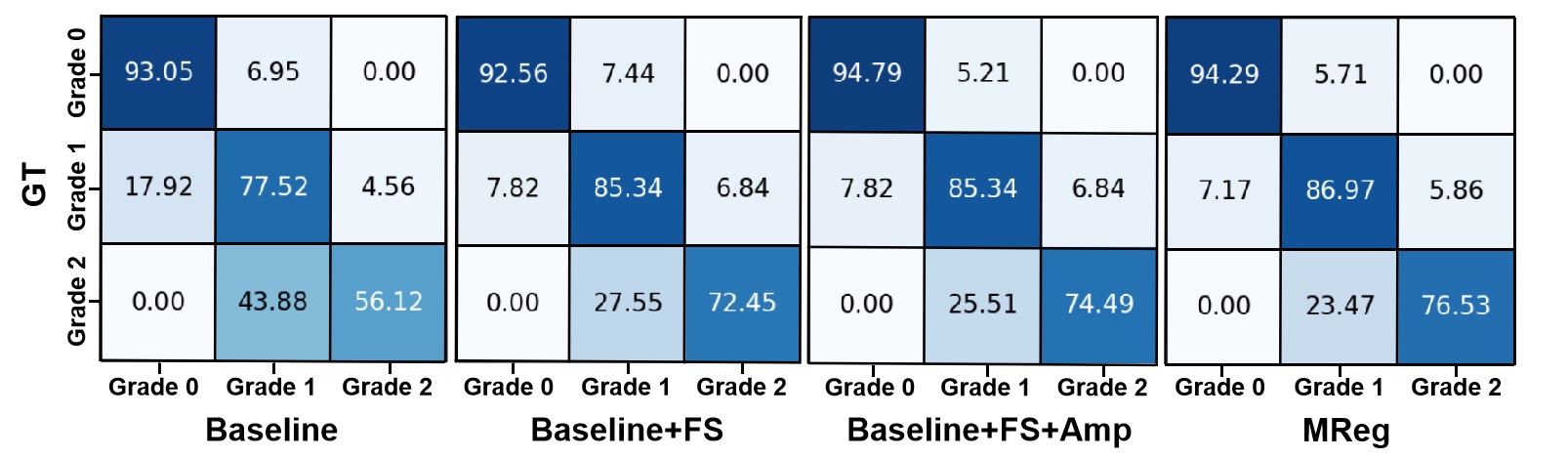}
\caption{Recall confusion matrix of the proposed method.
}
\label{fig:conf_result}
\end{figure}

\begin{figure}[!h]
\centering
\includegraphics[width=1.0\linewidth]{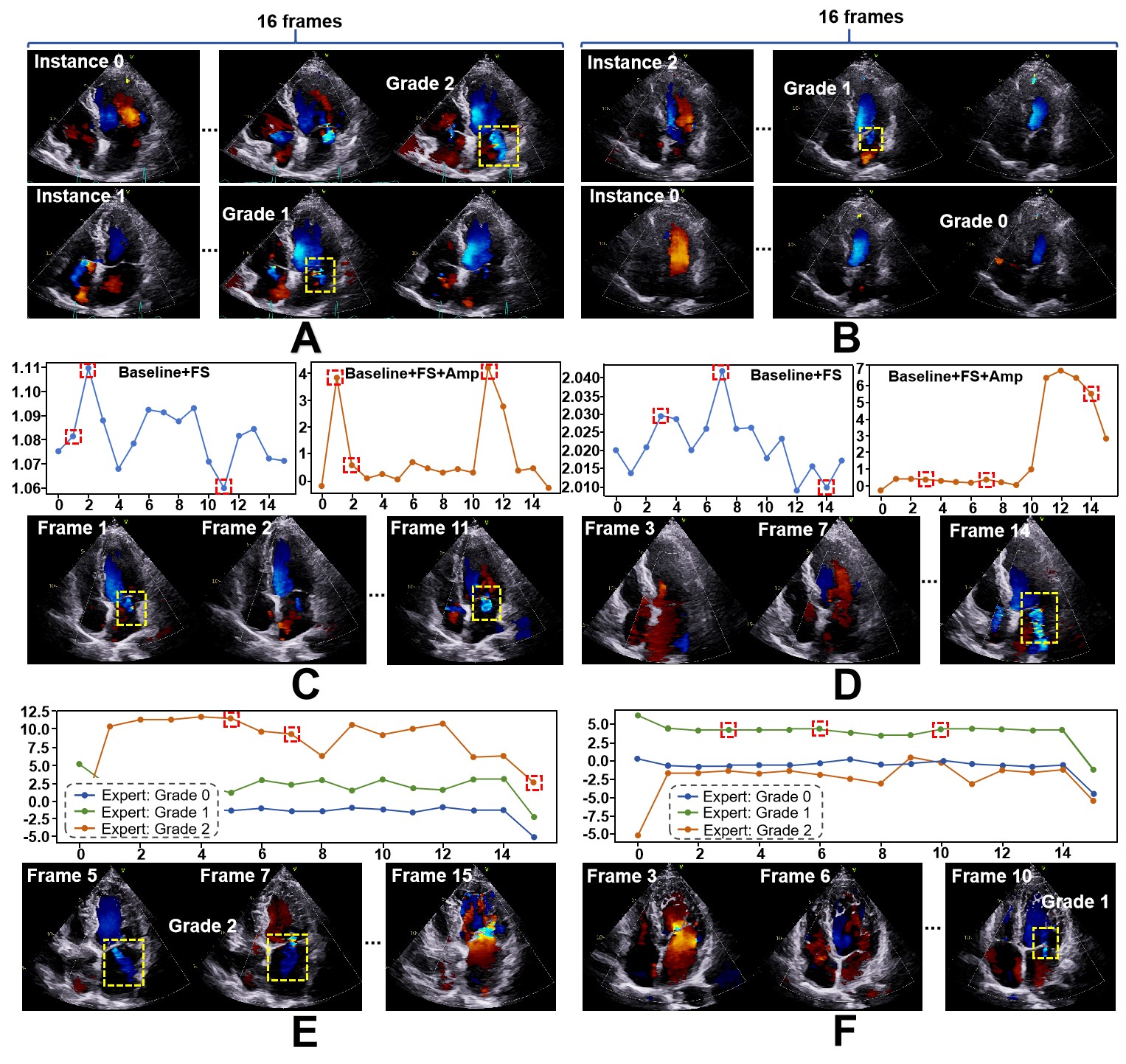}
\caption{\textbf{A} and \textbf{B} present instance selection for $MReg$ (row 1) and $MReg$** (row 2) based diagnosis, with the index on the first image.
\textbf{C}-\textbf{F} display per-frame regression results, including cases $w$ and $w/o$ Amp (C-D), and using feature summary (E-F).
Instance labels are marked on the corresponding frames.
} \label{fig:visi_result}
\end{figure}

Table~\ref{tab:ablation experiment} reports the ablation studies for different proposed modules, while Figure~\ref{fig:visi_result} visualizes the corresponding results.
We employ the X-CLIP-based single-stage regression model as the baseline. 
FS refers to the feature selection strategy, Amp denotes the feature amplification mechanism, MoE presents the feature summarization module, and $MReg$ signifies the final diagnostic model incorporating the loss $L_{expert}$. Results validate the effectiveness of regression-based MR diagnosis compared to classification-based solution, particularly for MR samples (rows 3-4, Table~\ref{tab:ablation experiment}, Recall: 82.99$\rightarrow$84.87).
Both FS and Amp significantly enhance diagnostic performance. 
The comparison between $MReg$** and $MReg$ further reveals the importance of accurate MR instance selection. 
Furthermore, as shown in Figure~\ref{fig:conf_result}, our method has a stable improvement in both Grades 1-2 with less data, especially in Grade 2 (56.12$\rightarrow$76.53, Baseline $vs.$ MReg).
While for Grade 0 with most data, MReg improves Baseline, and only degrades slightly compared to Baseline+FS+AMP.

\begin{figure}[!t]
\centering
\includegraphics[width=1.0\linewidth]{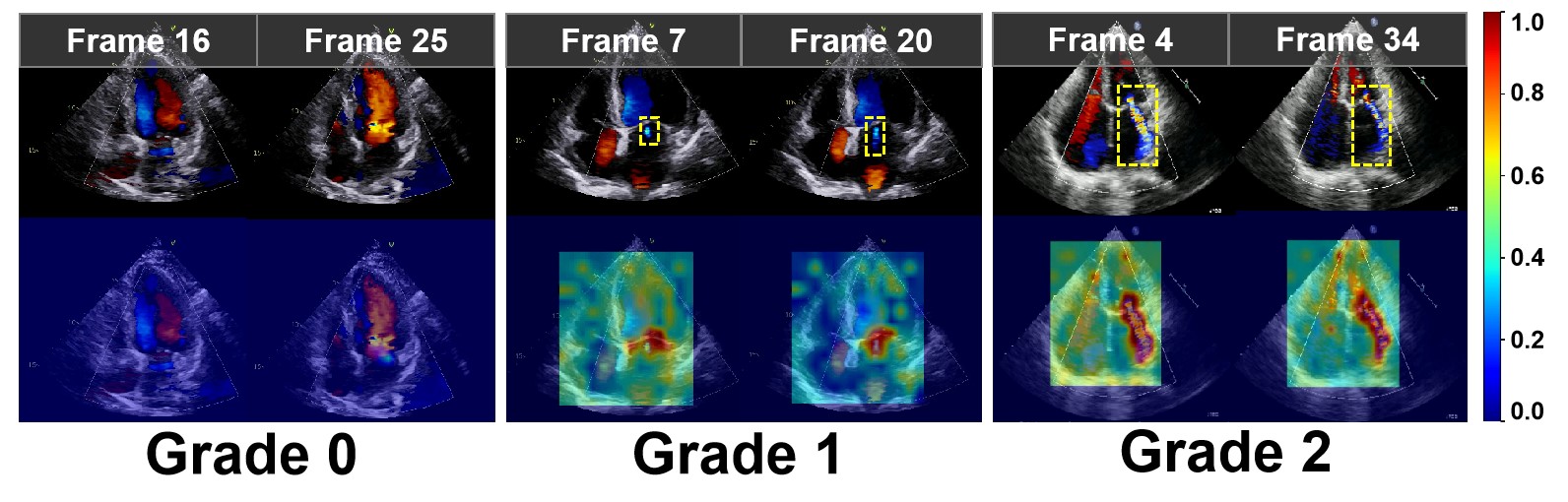}
\caption{The activation diagram of $MReg$ in the three categories, with the original image at the top and the corresponding activation maps at the bottom.} \label{fig:act_result}
\end{figure}

It can be seen in Figure~\ref{fig:visi_result} (A)-(B) that both $MReg$** cases resulted in incorrect judgments due to wrong instance selection.
For example, in Figure~\ref{fig:visi_result} (A), we provide two instances of the same video.
$MReg$ accurately finds instance 0 with the most severe regurgitation and predicts the right category, i.e., Grade 2.
However, $MReg$** selects instance 1 lacking key regurgitation information, thus making the wrong decision (Grade 1).
The results in Figures~\ref{fig:visi_result} (C) and (D) demonstrate that the regurgitation frame becomes more prominent after incorporating feature amplification (Amp), leading to more accurate predictions (as in Frames 1 and 11 in C).
Without Amp, Baseline+FS assigns low scores to key abnormal frames, resulting in prediction errors.  
Furthermore, the gap between decision frames and others in the same instance increases after applying Amp (as largest gaps: $\sim$0.05 $vs.$ $\sim$4 in C), making the decision frames dominant in the classification of the instance, reducing the influence of irrelevant frames.

Adding MoE improves the detection of positive samples, however, some metrics slightly decline (Accuracy/F1, Table~\ref{tab:ablation experiment}). 
We attribute this to the absence of a supervision mechanism to guide the learning direction of the added expert models. 
This can be supported by the improved performance observed with $L_{expert}$. Figure~\ref{fig:visi_result} (E) and (F) display the regression results for per frame in two instances after applying the feature summary module. 
The results reveal that the regression value of the expert model for the corresponding category is higher, indicating that different experts effectively learn distinct categories, consistent with our design assumption.
The activation maps in Figure~\ref{fig:act_result} further validate that our $MReg$ can accurately capture the vital regurgitation information.
Furthermore, both sets of instances in Figure~\ref{fig:visi_result} (C)-(F) show that the regression values increase with the severity of MR. 
The increased gap between different MR categories after applying Amp facilitates more precise MR differentiation. 

\section{Conclusion}
We propose a novel A4C-CDV-based MR diagnostic model, $MReg$ , to categorize normal, mild, and moderate-severe MR. 
We first design feature selection and amplification strategies to mimic the diagnostic logic of sonographers. 
Then, we propose a MoE-based feature summary module that enables different experts to summarize feature information in different categories for more accurate diagnosis. 
Experiments on our in-house dataset validate the effectiveness of our approach.
In the future, we will expand each MR subcategory to improve performance, and develop a multi-view diagnostic method.

\begin{credits}
\subsubsection{Acknowledgement.}
This work was supported by the National Natural Science Foundation of China (Nos. 82201851, 12326619, 62171290), Guangxi Province Science Program (No. 2024AB17023), Yunnan Major Science and Technology Special Project Program (No. 202402AA310052), Yunnan Key Research and Development Program 
(No. 202503AP140014), Suzhou Gusu Health Talent Program (Nos. GSWS2022071, GSWS2022072), Suzhou Key Laboratory of Cardiovascular Disease (No. SZS2024015) and Science and Technology Planning Project of Guangdong Province (No. 2023A05050 20002).

\subsubsection{Disclosure of Interests.}
The authors have no competing interests to declare that are relevant to the content of this article.
\end{credits}

\bibliographystyle{splncs04}
\bibliography{Paper-2951}

\end{document}